%% file: iccv17_OPN.tex
\documentclass[10pt,twocolumn,letterpaper]{article}

\usepackage{iccv}
\usepackage{times}
\usepackage{epsfig}
\usepackage{graphicx}
\usepackage{amsmath}
\usepackage{amssymb}
\usepackage{nth}
\usepackage{url}

\usepackage{tabularx}
\usepackage{paralist}
\usepackage{subfig}
\usepackage{booktabs}
\usepackage{multirow}

\usepackage[pagebackref=true,breaklinks=true,letterpaper=true,colorlinks,citecolor=blue,linkcolor=blue,bookmarks=false]{hyperref}

\include{macro}

\iccvfinalcopy 

\def\iccvPaperID{315} 

\begin{document}

\title{Unsupervised Representation Learning by Sorting Sequences}

\author{
Hsin-Ying Lee$^1$
\hspace{25pt} 
Jia-Bin Huang$^2$
\hspace{25pt} 
Maneesh Singh$^3$
\hspace{25pt} 
Ming-Hsuan Yang$^1$ \vspace{2pt}\\ 
$^1$University of California, Merced 
\hspace{25pt} 
$^2$Virginia Tech 
\hspace{25pt}
$^3$Verisk Analytics
\\
\url{http://vllab1.ucmerced.edu/~hylee/OPN/}
}

\maketitle

\begin{abstract}
We present an unsupervised representation learning approach using videos without semantic labels.
We leverage the temporal coherence as a supervisory signal by formulating representation learning as a sequence sorting task.
We take temporally shuffled frames (\ie in non-chronological order) as inputs and train a convolutional neural network to sort the shuffled sequences.
Similar to comparison-based sorting algorithms, we propose to extract features from all frame pairs and aggregate them to predict the correct order.
As sorting shuffled image sequence requires an understanding of the statistical temporal structure of images, training with such a proxy task allows us to learn rich and generalizable visual representation.
We validate the effectiveness of the learned representation using our method as pre-training on high-level recognition problems. 
The experimental results show that our method compares favorably against state-of-the-art methods on action recognition, image classification and object detection tasks.
%
%
\end{abstract}

\vspace{\secmargin}
\section{Introduction}
\label{sec:introduction}

In recent years, Convolutional Neural Networks (CNNs)~\cite{krizhevsky2012imagenet} have demonstrated the state-of-the-art performance in visual recognition tasks. 
The success of CNNs is primarily driven by millions of manually annotated data such as the ImageNet~\cite{deng2009imagenet}.
However, this substantially limits the scalability to new problem domains because manual annotations are often expensive and in some cases scarce (\eg labeling medical images requires expertise).
In contrast, a vast amount of \emph{free} unlabeled images and videos are readily available. 
It is of great interest to explore strategies for representation learning by leveraging unlabeled data.

A new unsupervised learning paradigm has recently emerged as \emph{self-supervised learning} \cite{raina2007self,wang2015unsupervised}.
Within the context of deep neural networks, the key idea is to leverage the inherent structure of raw images and formulate a discriminative or reconstruction loss function to train the network.
Examples include predicting the relative patch positions~\cite{doersch2015context}, reconstructing missing pixel values conditioned on the known surrounding~\cite{pathak2016inpainting}, or predicting one subset of the data channels from another (\eg predicting color channels from a gray image) ~\cite{larsson2016colorization,zhang2016colorization2,zhang2016splitbrain}.
Compared to image data, videos potentially provide much richer information as they not only consist of large amounts of image samples but also provide scene dynamics.
Recent approaches explore unlabeled video data to learn feature representation through egomotion~\cite{agrawal2015ego,jayaraman2015ego}, order verification~\cite{misra2016shuffle,fernando2016O3N}, tracking~\cite{wang2015unsupervised}, and future frame prediction~\cite{lotter2017Prenet}.
While these surrogate tasks do not directly use semantic labels, they provide effective supervisory signals as solving these tasks requires the semantic understanding of the visual data.

\begin{figure}[t]
\centering
\subfloat{%
\includegraphics[width=\linewidth]{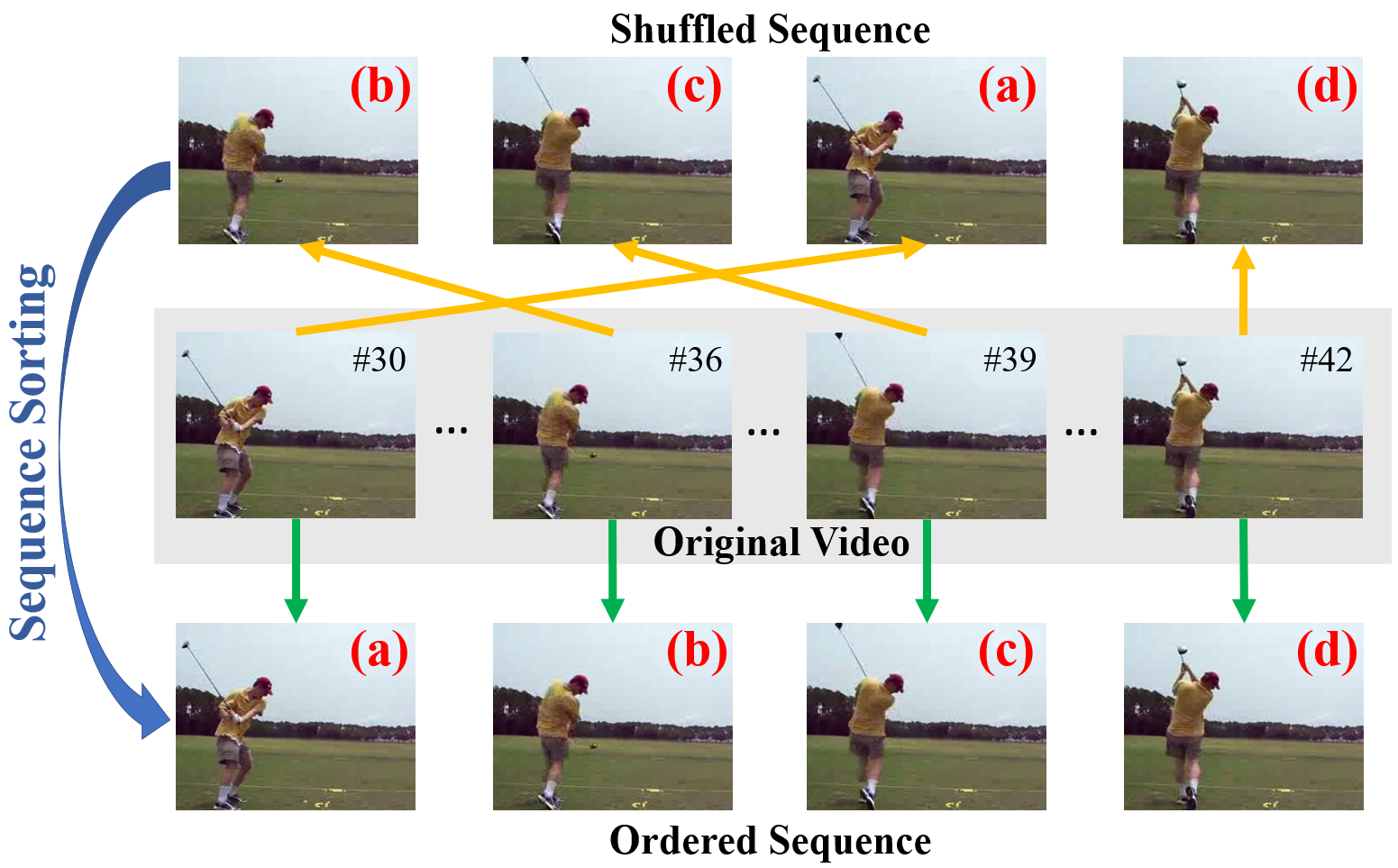}%
}
\caption{
\textbf{Representations learning by sequence sorting.}
Appearance variations and temporal coherence in videos offer rich supervisory signals for representation learning.
Given a tuple of randomly shuffled frames (top row) sampled from an input video (middle row), we formulate the sequence sorting task as revealing the underlying chronology order of the sampled frames (bottom row).
While no semantic labels are involved, solving this task requires high-level understanding of the temporal dynamics of videos.
In this paper, we exploit the statistical temporal structure of images as our source of supervision.
}
\label{figure:Overview}
\vspace{\figmargin}
\end{figure}

\begin{figure}[t]
\centering
\includegraphics[width=\linewidth]{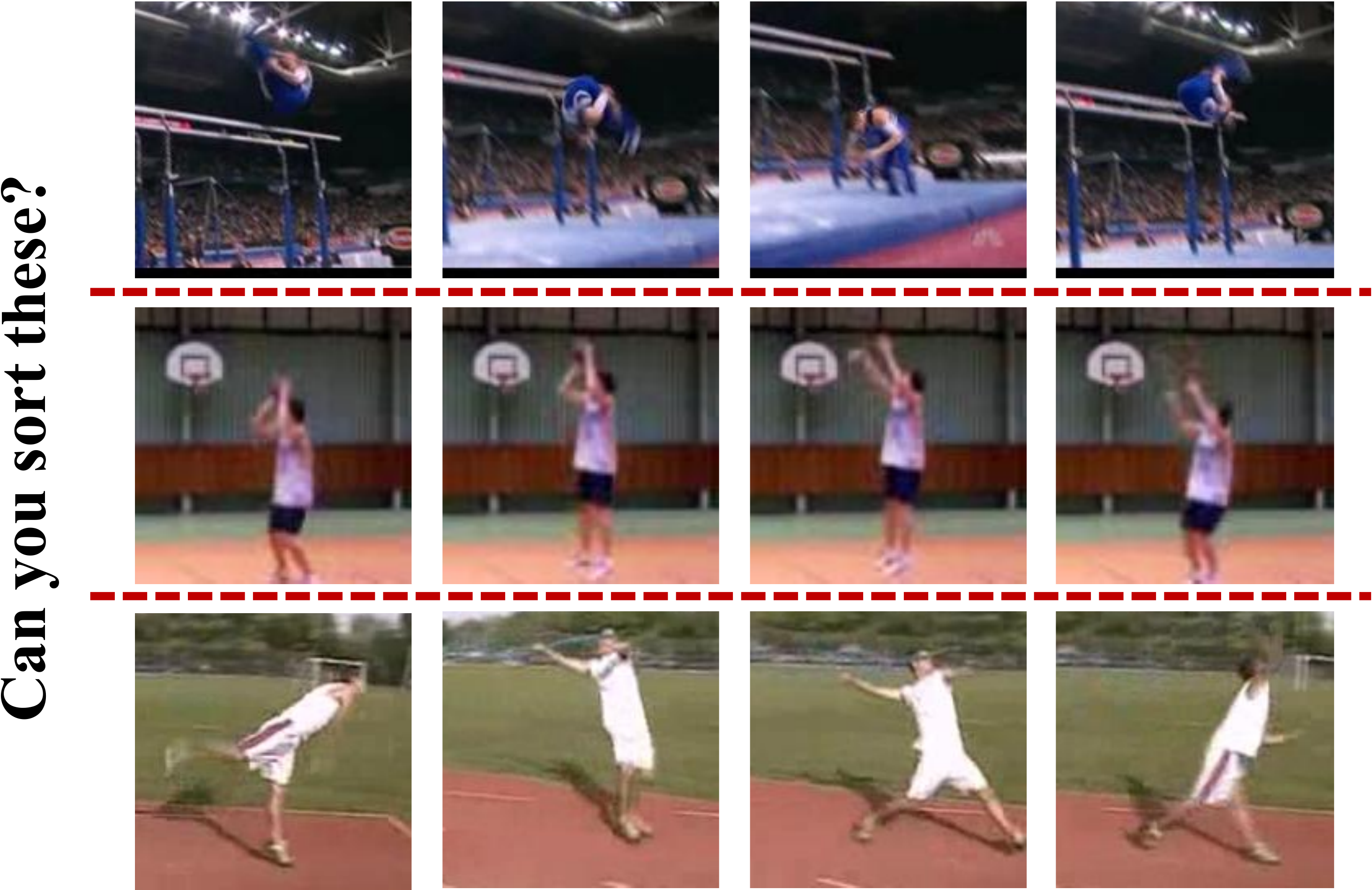}%
\vspace{-2mm}
\caption{
\textbf{Example tuples.} The examples shown here are automatically extracted tuples. Can you sort these shuffled frames? The answer is yes. By reasoning the relative poses using the knowledge of ``how a person moves", we can predict the chronological order of the frames.
}
\label{figure:Overview2}
\vspace{\figmargin}
\end{figure}

In this paper, we propose a surrogate task for self-supervised learning using a large collection of unlabeled videos.
Given a tuple of randomly shuffled frames, we train a neural network to sort the images into chronological order, as shown in \figref{Overview}.
The \emph{sequence sorting} problem provides strong supervisory signals as the network needs to reason and understand the statistical temporal structure of image sequences.
We show several examples of shuffled frames in~\figref{Overview2}. 
Our key observation is that we often unconsciously compare all pairs of frames to reason the chronological order of the sequence (as in comparison-based sorting methods).
In light of this, we propose an Order Prediction Network (OPN) architecture.
Instead of extracting features from all frames in a tuple simultaneously, our network first computes features from all the pairwise frames and fuses them for order prediction. 

We conduct extensive experimental validation to demonstrate the effectiveness of using sequence sorting for representation learning. 
When used as a pre-training module, our method outperforms state-of-the-art approaches on the UCF-101~\cite{soomro2012ucf101} and HMDB-51~\cite{kuehne2011hmdb} action benchmark datasets.
While our model learns features from human action videos, we also demonstrate the generalizability for generic object classification and detection tasks, and show competitive performance on the PASCAL VOC 2007 dataset~\cite{everingham2010pascal} when compared with the state-of-the-arts.


We make the following contributions in this work:

1) We introduce sequence sorting as a self-supervised representation learning approach using unlabeled videos.
While feature learning based on sequence order has been exploited recently~\cite{misra2016shuffle,fernando2016O3N,jayaraman2015slowandsteady}, 
our sorting formulation is much richer than the binary verification counterparts~\cite{misra2016shuffle}.

2) We propose an Order Prediction Network architecture to solve the sequence sorting task by pairwise feature extraction. Quantitative results show that the proposed architecture provides significant performance improvement over the straightforward implementation.

3) We show that the learned representation can serve as a pre-trained model.
Using less than 30,000 videos for unsupervised training, our model performs favorably against existing methods in action recognition benchmark datasets, and achieve competitive performance in classification and detection on the PASCAL VOC 2007 dataset.

\vspace{\secmargin}
\section{Related Work}
\label{sec:related work}

\paragraph{Unsupervised learning from static images.} 
While CNNs have shown dominant performance in high-level recognition problems such as classification and detection, training a deep network often requires millions of manually labeled images.
The inherent limitation from the fully supervised training paradigm highlights the importance of unsupervised learning to leverage vast amounts of unlabeled data.
Unsupervised learning has been extensively studied over the past decades. 
Before the resurgence of CNNs, hand-craft features such as SIFT and HOG have been used 
to discover semantic classes using clustering~\cite{russell2006handcraft1,sivic2005handcraft2}, 
or mining discriminative mid-level features~\cite{singh2012midlevel1,doersch2013midlevel2,sun2013midlevel3}.
With deep learning techniques, rich visual representations can be learned and extracted directly from images. 
A large body of literature focuses on reconstruction-based learning.
Inspired from the original single-layer auto-encoders~\cite{olshausen1997oldautoencoder}, several variants have been developed, including stack layer-by-layer restricted Boltzmann machines (RBMs), and auto encoders~\cite{bengio2007stackingRBMandAutoencoder,
hinton2006stackingRBM,le2013autoendocerLargescale}.

Another line of unsupervised learning is known as self-supervised learning. 
These methods define a supervisory signal for learning using the structure of the raw visual data. 
The spatial context in an image provides a rich source of supervision.
Various existing approaches leverage spatial context for self-supervision, including predicting the relative patch positions~\cite{doersch2015context}, solving jigsaw puzzles~\cite{noroozi2016puzzle}, and inpainting missing regions based on their surrounding~\cite{pathak2016inpainting}.
Another type of cue is through cross-channel prediction, \eg image colorization~ \cite{larsson2016colorization,zhang2016colorization2} and split-brain auto-encoders~\cite{zhang2016splitbrain}. 
In addition to using only individual images, several recent directions have been explored by grouping visual entities using co-occurrence in space and time~\cite{isola2015cooccur}, using graph-based constraints~\cite{li2016unsupervised}, and cross-modal supervision from sounds~\cite{owens2016ambient}.
Our work is similar to context-based approaches~\cite{doersch2015context,noroozi2016puzzle,pathak2016inpainting}. 
Instead of using \emph{spatial} context of images, in this work we investigate the use of \emph{temporal} context in videos.

\begin{figure*}[t]
\centering
\includegraphics[width=.85\linewidth]{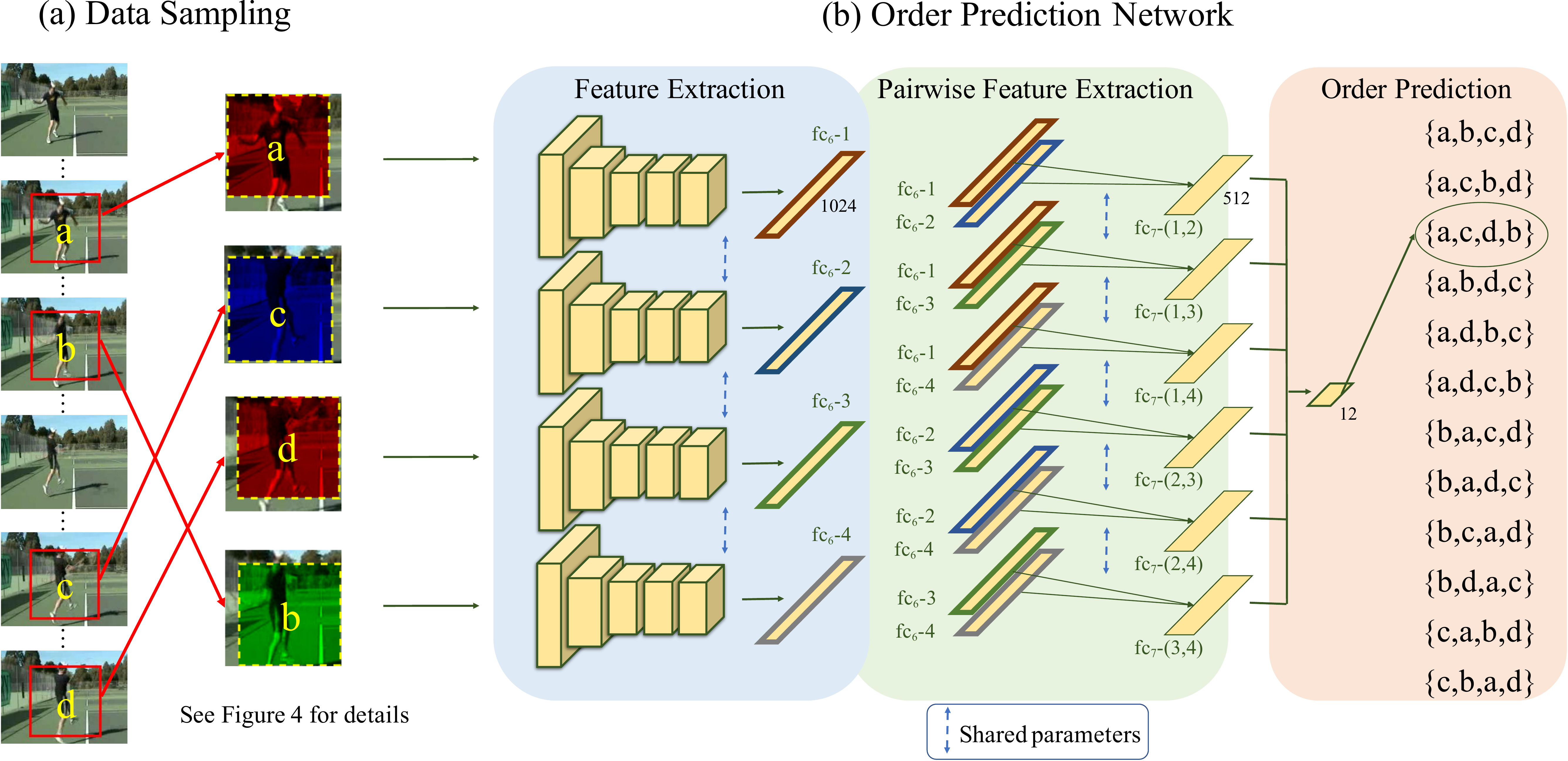}
\caption{\textbf{Overview of our approach.} 
Our training strategy consists of two main steps. 
(a) \emph{Data sampling} (Section~\ref{subsec:datasample}). 
We sample candidate tuples from an input video based on motion magnitude.
We then apply spatial jittering and channel splitting on selected patches to guide the network to focus on the semantics of the images rather than fixating on low-level features. 
Finally, we randomly shuffle the sampled patches to form an input tuple for training the CNN.
(b) \emph{Order Prediction Network} (Section~\ref{subsec:architecture}). 
The proposed Order Prediction Network consists of three main components: (1) feature extraction, (2) pairwise feature extraction, and (3) order prediction. 
Features for each frame ($fc_6$) are encoded by convolutional layers. 
The pairwise feature extraction stage then extracts features from every pair of frames.
We then have a final layer that takes these extracted features to predict order.
Note that while we describe the architecture in three separate stages for the sake of presentation, the network training is end-to-end without stage-wise optimization.}
\label{figure:architecture}
\vspace{\figmargin}
\end{figure*}

\vspace{\paramargin}
\paragraph{Unsupervised learning from videos.}
The explosive increase of easily available videos on the web, like YouTube, presents an opportunity as well as several challenges for learning visual representations from unlabeled videos.
Compared to images, videos provide the advantage of having an additional time dimension.
Videos provide examples of appearance variations of objects over time.
We can broadly categorize the unsupervised learning methods using videos into two groups.
The first group focuses on frame reconstruction tasks, \eg future frame prediction~\cite{srivastava2015LSTM}, frame interpolation~\cite{long2016interpolation}, and video generation~\cite{vondrick2016generating}.
The second group learns feature representation by leveraging appearance variations presented in videos.
Examples include enforcing the temporal smoothness of representation throughout a video~\cite{mobahi2009temporalcoherence,jayaraman2015slowandsteady}, applying tracking to capture appearance variation of moving objects~\cite{wang2015unsupervised}, 
learning transformation in ego-motion videos~\cite{jayaraman2015ego,agrawal2015ego}, verifying the order of input sequence~\cite{misra2016shuffle,fernando2016O3N}, and the transformation between color and optical flow~\cite{purushwalkam2016transformation}. 

The work most related to our method is that of~\cite{misra2016shuffle,fernando2016O3N}. 
Similar to Misra~\etal~\cite{misra2016shuffle}, our method makes use of the temporal order of frames as the source of supervision.
However, instead of verifying correct/incorrect temporal order (\ie binary classification), our supervisory signals are much richer: our network needs to predict $n!/2$ combinations for each n-tuple of frames.
The proposed Order Prediction Network architecture also differs from the simple concatenation in~\cite{misra2016shuffle}.
The Order Prediction Network first extracts pairwise features and subsequently fuse the information for final predictions.
Our quantitative results demonstrate performance improvement using the proposed design.
Fernando~\etal~\cite{fernando2016O3N} exploit a similar notion of order verification to learn video representation.
However, their approach takes as input a stack of frame differences and does not learn image representations.
In contrast, our model can be used for \emph{both} video understanding (\eg action recognition) as well as image understanding (\eg classification and detection) problems (as we show in \secref{experiments}).

\vspace{\secmargin}
\section{Feature Learning by Sequence Sorting}
\label{sec:overview}

Our goal is to capitalize the large quantity of \emph{unlabeled} videos for feature learning.
We propose to use sequence sorting as a surrogate task for training a CNN.
Our hypothesis is that successfully solving the sequence sorting task will allow the CNN to learn useful visual representation to recover the temporal coherence of video by observing how objects move in the scene. 

Specifically, we use up to four randomly shuffled frames sampled from a video as our input.
Similar to the jigsaw puzzle problem in the spatial domain~\cite{noroozi2016puzzle}, we formulate the sequence sorting problem as a multi-class classification task.
For each tuple of four frames, there are $4 != 24 $ possible permutations.
However, as some actions are both coherent forward and backward (\eg opening/closing a door), we group both forward and backward permutations into the same class (\eg $24/2$ classes for four frames).
This forward-backward grouping is conceptually similar to the commonly used horizontal flipping for images.
In the following, we describe two important factors in our approach: (1) training data sampling (\subsecref{datasample}) and (2) network architecture (\subsecref{architecture}).

\vspace{\secmargin}
\subsection{Training data sampling}
\label{subsec:datasample}

Preparing training data is crucial for self-supervised representation learning.
In the proposed sequence sorting task, we need to balance the level of difficulty.
On the one hand, sampling tuples from static regions produces nearly impossible tasks for the network to sort the shuffled sequence.
On the other hand, we need to avoid the network picking up low-level cues to achieve the task.
We describe three main strategies to generate our training data in this section.

\begin{figure}[t]
\centering
\subfloat[Motion-aware tuple selection]{%
\includegraphics[width=\columnwidth]{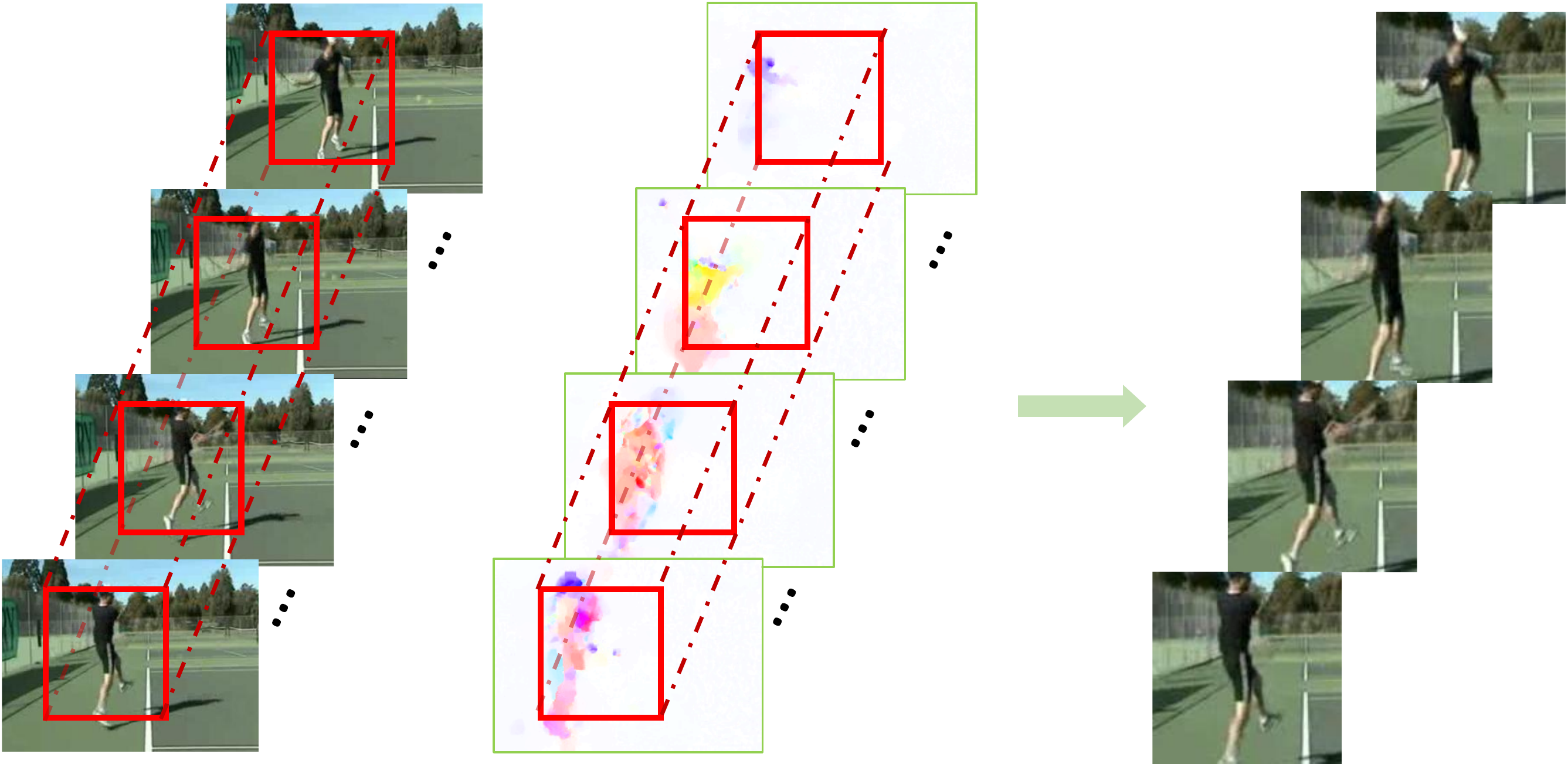}%
}
\vspace{\subfigmargin}
\subfloat[Spatial jittering]{%
\includegraphics[width=\columnwidth]{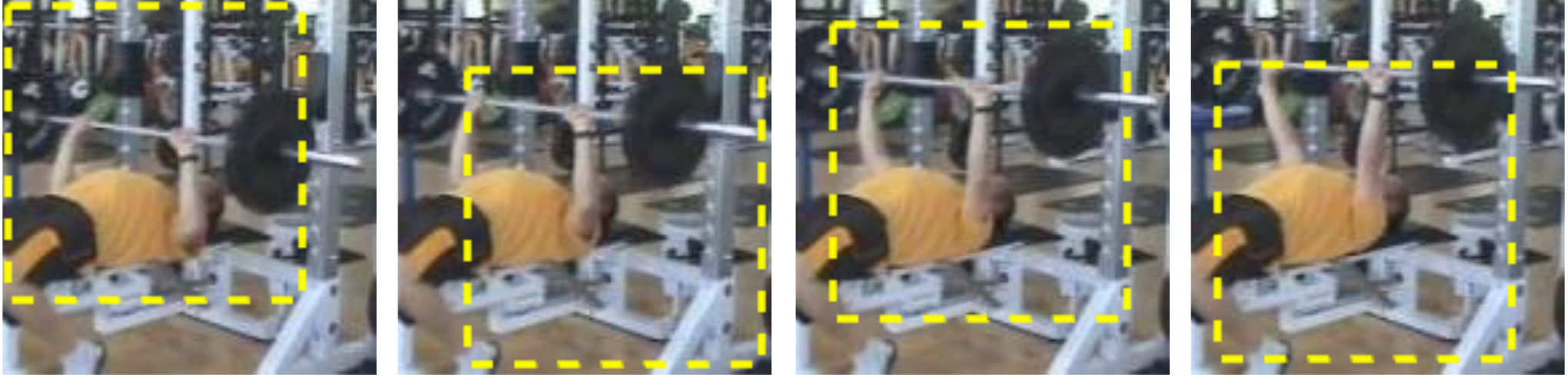}%
}
\vspace{\subfigmargin}
\subfloat[Channel splitting]{%
\includegraphics[width=\columnwidth]{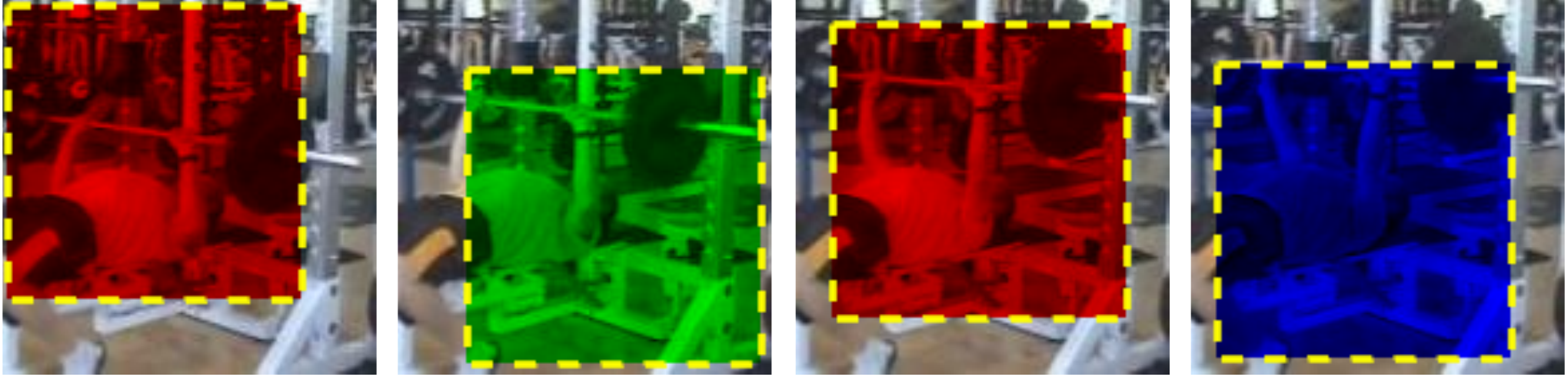}%
}
\caption{\textbf{Data sampling strategies.} 
(a) We use a sliding windows approach on the optical flow fields to extract patches tuple with large motion magnitude. 
(b)(c) To prevent the network from extracting low-level features for sorting, we apply random spatial jittering and channel splitting as our data augmentation. 
Quantitative results show that these data augmentation strategies lead to improved performance.
}
\label{figure:datasampling}
\vspace{\figmargin}
\end{figure}

\vspace{\paramargin}
\paragraph{Motion-aware tuple selection.}
We use the magnitude of optical flow to select frames with large motion regions similar to~\cite{misra2016shuffle}.
In addition to using optical flow magnitude for frame selection, we further select spatial patches with large motion.
Specifically, for video frames in the range $[t_{min}, t_{max}]$, we use sliding windows to mine frame tuple $\{t_a,t_b,t_c,t_c\}$ with large motion, as illustrated in \figref{datasampling}(a).

\vspace{\paramargin}
\paragraph{Spatial jittering.} 
As the previously selected tuples are extracted from the same spatial location, simple frame alignment could potentially be used to sort the sequence. 
We apply spatial jittering for each extracted patch to avoid the trivial cases (see \figref{datasampling}(b)). 

\vspace{\paramargin}
\paragraph{Channel splitting.} 
To avoid the network from learning low-level features without semantic understanding, we apply channel splitting on the selected patches, as shown \figref{datasampling}(c).
For each frame in a tuple, we randomly choose one channel and duplicate the values to other two channels.
The effect is similar to using a grayscale image (as done in~\cite{wang2015unsupervised}). 
However, the use of channel splitting imposes additional challenges for the network compared with using grayscale images because grayscale images are generated from a fixed linear combination of the three color channels.
We validate all design choices in \subsecref{ablation}.

\vspace{\secmargin}
\subsection{Order Prediction Network}
\label{subsec:architecture}
The proposed OPN has three main components: 
(1) frame feature extraction, 
(2) pairwise feature extraction, 
(3) order prediction. 
\figref{architecture} shows the architecture in the case of 4-tuple.

\vspace{\paramargin}
\paragraph{Frame feature extraction.}
Features for each frame ($fc6$) are encoded by convolutional layers. 
We use a Siamese architecture where all the branches share the same parameters.

\vspace{\paramargin}
\paragraph{Pairwise feature extraction.}
A straightforward architecture design for solving the order prediction problem is to concatenate either $fc6$ or $fc7$ features for the frames and use the concatenation as the representation of the input tuple.
However, such ``taking one glimpse at all frames'' approach may not capture the concept of \emph{ordering} well.
Therefore, inspired from comparison-based sorting algorithms, we propose to perform pairwise feature extractions on extracted features.
Specifically, we take the $fc6$ features from every pair of frames for extractions. 
For example, in \figref{architecture}, the layer$7$-$(1,2)$ provides information of the relationship of the first and second frames.

\vspace{\paramargin}
\paragraph{Order prediction.}
The final order prediction is then based on the concatenation of all pairwise feature extractions after one fully connected layer and softmax function.

\vspace{\secmargin}
\subsection{Implementation details}
\label{subsec:implementation}
We implement our method and conduct all experiments using the Caffe toolbox~\cite{jia2014caffe}.
We use the CaffeNet~\cite{jia2014caffe}, a slight modification of AlexNet~\cite{krizhevsky2012imagenet}, as our architecture for convolutional layers. 
For the sake of efficiency, our network takes $80 \times 80$ patches as inputs. 
It dramatically reduces the number of parameters and training time. 
Our network has only $5.8$M parameters up to $fc7$, compared to the $58.2$M parameters used in $AlexNet$. 
As the architecture using feature concatenation have $9$M parameters, the performance gain of OPN does not come from the number of parameters.

We use stochastic gradient descent with a momentum of $0.9$ and a dropout rate of $0.5$ on fully connected layers. 
We also use batch normalization~\cite{ioffe2015batchnorm} on all layers. 
We extract 280k tuples from the UCF-101 dataset as our training data. 
To train the network, we set the batch size as $128$ and the initial learning rate as $10^{-2}$.
We reduce the learning rate by a factor of 10 at 130k and 350k iterations, with a total of 200k iterations.
The entire training process takes about 40 hours on one Titan X GPU.
All the pre-trained models and the source code are available in the project page. \footnote{\url{http://vllab1.ucmerced.edu/~hylee/OPN/}}

\vspace{\secmargin}
\section{Experiments}
\label{sec:experiments}

In this section, we validate the effectiveness of the learned representation.
First, we treat our method as an unsupervised pre-training approach to initialize models for action recognition (\subsecref{actionrecog}), image classification, and object detection (\subsecref{pascalvoc}).
Second, we conduct an ablation study to quantify the contributions from individual components of our approach (\subsecref{ablation}).
Third, we visualize the low-level filters and high-level activations (\subsecref{vis}). 
%
%
Below we describe the variants of our model:
\begin{compactitem}
\item \tb{binary: } Order verification similar to~\cite{misra2016shuffle}.
\item \tb{3-tuple:} Takes a tuple of 3 frames as input and predicts $3!/2=3$ classes. 
\item \tb{4-tuple:} Take a tuple of 4 frames as input and predicts $4!/2=12$ classes. 
\item \tb{Concat:} Prediction order from the concatenation of $fc6$ features after two fully connected layers.
\end{compactitem}

\vspace{\secmargin}
\subsection{Action recognition}
\label{subsec:actionrecog}
We use our approach as a pre-training method on the action recognition datasets. 
We compare our model with Misra \etal~\cite{misra2016shuffle} and Fernando \etal~\cite{fernando2016O3N} which learn features by verifying the order correctness, 
Purushwalkam \etal~\cite{purushwalkam2016transformation} which views optical flows features as transformation between RGB features, and Vondrick \etal~\cite{vondrick2016generating} which applies GAN to generate videos.

\vspace{\paramargin}
\paragraph{Datasets.}
We use the three splits of the UCF-101~\cite{soomro2012ucf101} and HMDB-51~\cite{kuehne2011hmdb} action recognition datasets to evaluate the performance of our unsupervised pre-trained network. 
The UCF-101 dataset consists of 101 action categories with about 9.5k videos for training and 3.5k videos for testing.
The HMDB-51 dataset consists of 51 action categories with about 3.4k videos for training and 1.4k videos for testing. 
We evaluate the classification accuracy on both datasets.

\begin{table}[t]
\caption{\textbf{Mean classification accuracy over the three splits of the UCF-101 dataset}. $^*$Purushwalkam \etal~\cite{purushwalkam2016transformation} use videos from the UCF-101 (split 1), HMDB-51 (split 1), and ACT datasets for training. $^{\dag}$Vondrick \etal~\cite{vondrick2016generating} use C3D as their architecture. They use videos downloaded from Flickr.}
\label{tab:actionRecogUCF}
\vspace{\tabmargin}
\centering
\small
\begin{tabular}{l cc} \toprule
Initialization& CaffeNet & VGG-M-2048 \\
\midrule
random &47.8 & 51.1\\ 
ImageNet&67.7 & 70.8\\
\midrule
Misra \etal~\cite{misra2016shuffle}&50.2 & - \\
Purushwalkam \etal~\cite{purushwalkam2016transformation}*& - &55.4 \\
Vondrick \etal~\cite{vondrick2016generating}$^{\dag}$ & 52.1 & - \\ 
\hline
binary& 51.6 & 56.8\\
3-tuple Concat& 52.8 & 57.0\\
3-tuple OPN& 53.2 & 58.3 \\
4-tuple Concat& 55.2 &59.0 \\
4-tuple OPN& \textbf{56.3} & \textbf{59.8}\\
\bottomrule
\end{tabular}
\vspace{\tabmargin}
\end{table}

\begin{table}[t]
\caption{\textbf{Mean classification accuracy over the three splits of the HMDB-51 dataset}. Methods with ``(UCF)'' are pre-trained on the UCF-101 dataset. $^*$Purushwalkam \etal~\cite{purushwalkam2016transformation} uses videos from the UCF-101 (split 1), HMDB-51 (split 1), and ACT datasets for training.}
\label{tab:actionRecogHMDB}
\vspace{\tabmargin}
\centering
\small
\begin{tabular}{l cc} \toprule
Initialization & CaffeNet & VGG-M-2048 \\
\midrule
random & 16.3 & 18.3 \\ 
Imagenet & 28.0 & 35.3 \\
\midrule
Misra \etal~\cite{misra2016shuffle}&18.1 & - \\
Purushwalkam \etal~\cite{purushwalkam2016transformation}*& - &23.6 \\
\midrule
binary & 20.9 & 21.0 \\
3-tuple OPN & 21.3 & 21.5 \\
4-tuple OPN & 21.6 & 21.9\\
\midrule
Misra \etal~\cite{misra2016shuffle} (UCF) & 15.2 & -\\
4-tuple OPN (UCF) & \textbf{22.1} & \textbf{23.8}\\
\bottomrule
\end{tabular}
\vspace{\tabmargin}
\end{table}

\begin{table}[]
\caption{\textbf{Comparison with O3N}~\cite{fernando2016O3N}. The baseline is not the same because O3N uses stacks of frame differences (15 channels) as inputs. To use a similar setting, we take single frame difference (Diff) as inputs and initialize the weights with models trained on RGB and Diff features.}
\label{tab:O3N}
\centering
\small
\vspace{\tabmargin}
\begin{tabular}{l cccc} \toprule
Method & unsupervised & supervised& UCF & HMDB \\
\midrule
O3N~\cite{fernando2016O3N}&Stack of Diff& Stack of Diff & 60.3 &32.5 \\
\hline
OPN& RGB & Diff & \textbf{71.8} &36.7\\
OPN& Diff & Diff&71.4& \textbf{37.5}\\
\bottomrule
\end{tabular}
\vspace{\tabmargin}
\end{table}

\input{table_pascal}

\vspace{\paramargin}
\paragraph{Results.}

%
After training with \emph{unlabeled} videos from UCF-101, we fine-tune the model using the labeled videos.
\tabref{actionRecogUCF} and \tabref{actionRecogHMDB} shows the results on the UCF-101 and HMDB-51 datasets, respectively. 
Overall, the quantitative results show that more difficult tasks provide stronger semantic supervisory signals and guide the network to learn more meaningful features.
The OPN obtains 57.3\% accuracy compared to 52.1\% of from Vondrick~\etal~\cite{vondrick2016generating} on the UCF-101 dataset.
To compare with~\cite{purushwalkam2016transformation}, we also train our model using VGG-M-2048~\cite{simonyan2014VGG}.
Note that the method in~\cite{purushwalkam2016transformation} uses the UCF-101, HMDB-51 and ACT datasets to train their model (about 20k videos).
In contrast, our OPN uses videos from the UCF-101 training set and outperforms~\cite{purushwalkam2016transformation} by 5.1\%.
While using 3k videos from the HMDB-51 dataset for both unsupervised and supervised training, OPN performs slightly worse than~\cite{purushwalkam2016transformation}.

We also compare with a recent method for video representation learning~\cite{fernando2016O3N}. 
It is difficult to have a fair comparison because they use \emph{stacks of frame differences} (15 channels) as inputs rather than RGB images. 
To use a similar setting, we take single frame difference $Diff(t) = RGB(t+1)-RGB(t)$ as inputs to train our model.
We initialize the network with models trained on $RGB$ and $Diff$ features.
As shown in \tabref{O3N}, our method compares favorably against~\cite{fernando2016O3N} by more than 10\% gain on the UCF-101 dataset and 5\% on the HMDB-51 dataset. 
The performance of initializing with the model trained on $RGB$ features is similar to with model trained on frame difference. 
The results demonstrate the generalizability of our model.

We also evaluate the transferability of our learned features.
We initialize the weights with the model trained on the UCF-101 training set (without using any labels).
\tabref{actionRecogHMDB} shows the results.
Our method achieves 22.5\% compared to 15.2\% of~\cite{misra2016shuffle} under the same setting.
We achieve slightly higher performance when there is no domain gap (\ie using training videos from the HMDB-51 dataset).
The results suggest that our method is not heavily data dependent and is capable of learning generalizable representations.

\vspace{\secmargin}
\subsection{PASCAL VOC 2007 classification and detection}
\label{subsec:pascalvoc}

To evaluate the generalization ability, we use our model as pre-trained weights for classification and detection tasks. 
We compare our approach with recent image-based and video-based unsupervised learning methods.

\vspace{\paramargin}
\paragraph{Dataset.} 
The PASCAL VOC 2007~\cite{everingham2010pascal} dataset has 20 object classes and contains 5,011 images for training and 4,952 images for testing.
We train our model using the UCF-101, HMDB-51, and ACT~\cite{wang2016act} datasets.
For both tasks we use the same fine-tuning strategy described in Kr{\"a}henb{\"u}hl \etal~\cite{krahenbuhl2015datadependent} without the rescaling method.
We use the \textit{CaffeNet} architecture and the Fast-RCNN~\cite{girshick2015fastrcnn} pipeline for the detection task.
We evaluate all algorithms using the mean average precision (mAP)~\cite{everingham2010pascal}. 
Since our fully connected layers are different from the standard network, we copy only the weights of the convolutional layers and initialize the fully connected layers from a Gaussian distribution with mean $0$ and standard deviation $0.005$. 
For a fair comparison with existing work, we train and test our models without using batch normalization layers.

\vspace{\paramargin}
\paragraph{Results.}

\tabref{pascal} lists the summary of methods using static images and method using videos.
While our performance is competitive, methods trained with ImageNet performs better than that using videos.
We attribute this gap to the fact that the training images are object-centric while our training videos are human-centric (and thus may not contain diverse appearance variations of generic objects). 
Among the methods using videos, our method shows competitive performance to ~\cite{wang2015unsupervised}. 
However, our method requires considerably less training time and less number of training videos.
%



\vspace{\secmargin}
\subsection{Ablation analysis}
\label{subsec:ablation}

We evaluate the effect of various design choices on the split 1 of the UCF-101 dataset.
We first perform unsupervised pre-training using the videos from the training set. 
The learned weights are then used as the initialization for the supervised action recognition problem.

\vspace{\paramargin}
\paragraph{Motion.}
We select our training tuples according to the magnitude of optical flow. 
To demonstrate the necessity of this step, we compare it with randomly selecting frames from a video. 
We also use the optical flow direction as a further restriction.
Specifically, the motion in the selected interval must remain in the same direction.
\tabref{dataSampling_motion} shows the results of how these tuple selection methods affect the final performance.
Using random selection degrades the performance because the training data contain many similar patches that are impossible to be sorted (\eg static regions).
We also observe that adding the direction constraint does not help.
The direction constraint eliminates many tuples with shape deformation (\eg pitching contains motions in reverse direction).
The network thus is unable to learn meaningful high-level features.

\begin{table}[t]
\caption{
\textbf{Comparison of different sampling strategies.}
\textit{Motion} uses the magnitude of optical flow for patches selection.
\textit{Direction} further restricts the monotonicity of optical flow direction in selected tuples. The results show that \textit{Direction} oversimplifies the problems and thus degrades the performance.
}
\label{tab:dataSampling_motion}
\vspace{\tabmargin}
\centering
\begin{tabular}{lcc}\toprule
Strategy & Action Recognition (\%) \\
\midrule
Random& 47.2 \\
Motion& \textbf{57.3}\\
Motion+Direction& 52.6\\
\bottomrule
\end{tabular}
\vspace{\tabmargin}
\end{table}

\begin{table}[t]
\caption{\textbf{Comparison of using different patch sizes.} Using $80\times80$ patches has advantages in all aspects. }
\label{tab:dataSampling_patch}
\vspace{\tabmargin}
\centering
\small
\begin{tabular}{c c c c} \toprule
\multirow{2}{*}{Patch size} & \multirow{2}{*}{$\#$Parameters} & \multirow{2}{*}{Training time} & Action \\
& & & Recognition (\%)\\
\midrule
80& \textbf{5.8M}& \textbf{1x} & \textbf{57.3} \\
120& 7.1M& 1.4$\times$& 55.4\\
224& 14.2M& 2.2$\times$ & 51.9\\
\bottomrule
\end{tabular}
\vspace{\tabmargin}
\end{table}

\vspace{\paramargin}
\paragraph{Patch size.}
We experiment with different patch sizes for training the network. 
Due to the structure of fully connected layers, the patch size selection significantly affects the number of parameters and the training time.
\tabref{dataSampling_patch} shows the comparison among using patch size $80\times 80$, $120\times 120$, and the entire image. 
The results show that using $80\times 80$ patches has an advantage in terms of the number of parameters, training time, and most importantly, the performance. 
One potential reason for the poor performance of using larger patches might be the insufficient amount of video training data.

\vspace{\paramargin}
\paragraph{Spatial jittering.}
Analogous to the random gap used in the context prediction task~\cite{doersch2015context} and puzzle-solving task~\cite{noroozi2016puzzle}, we apply spatial jittering to frames in a tuple to prevent the network from learning low-level statistics. 
In practice, we apply random shift of $[-5, 5]$ pixels to bounding boxes in both horizontal and vertical directions.
\tabref{dataSampling_spacejit} shows the applying spatial jittering does further help the network to learn better features.

\begin{table}[t]
\caption{ \textbf{Effect of spatial jittering.} For both 3-tuple and 4-tuple cases, OPNs with spatial jittering perform better.}
\label{tab:dataSampling_spacejit}
\vspace{\tabmargin}
\centering
\small
\begin{tabular}{lc c} 
\toprule
Method & Spatial jittering & Action Recognition (\%) \\
\midrule
3-tuple OPN& & 55.8\\
3-tuple OPN& \checkmark &56.1 \\
4-tuple OPN& & 56.5\\
4-tuple OPN& \checkmark & 57.3\\
\bottomrule
\end{tabular}
\vspace{\tabmargin}
\end{table}

%
\begin{table}[t]
\caption{\textbf{Effect of pairwise feature extraction on order prediction and action recognition.} The results demonstrate the performance correlation between two tasks, and show that OPN facilitates the feature learning.}
\label{tab:dataSampling_suborder}
\vspace{\tabmargin}
\centering
\small
\begin{tabular}{lcc} \toprule
\multirow{2}{*}{Method} & Order & Action \\
& Prediction (\%) & Recognition (\%) \\
\midrule
3-tuple Concat& 59 & 53.4\\
3-tuple OPN& \textbf{63}& \textbf{54.1}\\
\midrule
4-tuple Concat& 38 & 56.1\\
4-tuple OPN& \textbf{41} & \textbf{57.3} \\
\bottomrule
\end{tabular}
\vspace{\tabmargin}
\end{table}

\begin{figure}[t]
\centering
\includegraphics[width=\columnwidth]{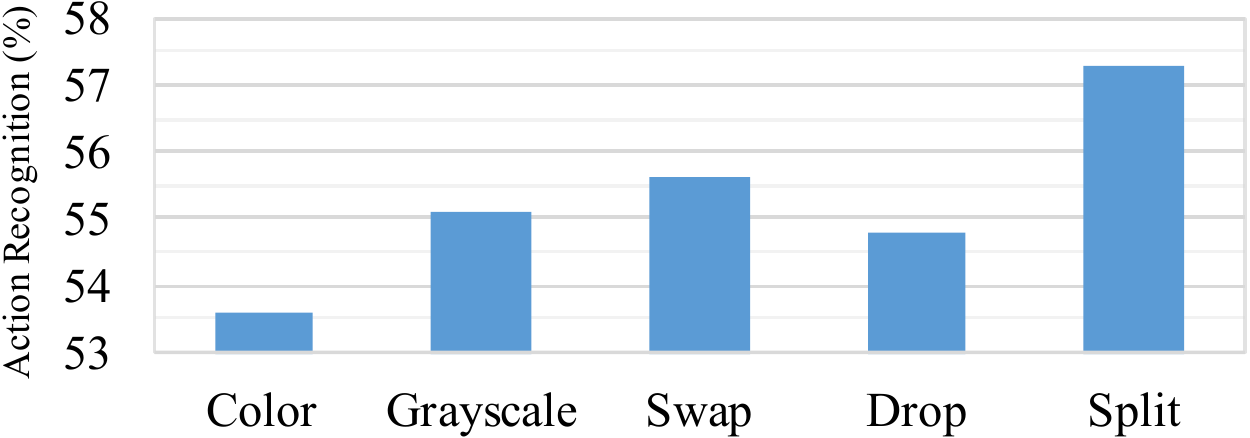}
\caption{\textbf{Effect of different strategies on color channels.} The proposed channel splitting outperforms other strategies.}
\label{figure:dataSampling_colorsplit}
\vspace{\figmargin}
\end{figure}

\vspace{\paramargin}
\paragraph{Channel splitting.}
To further prevent the network from learning trivial features, we reduce the visual clues from color.
The most intuitive way is to use the grayscale image. 
However, grayscale images are generated from a fixed linear combination of the three color channels. 
To mitigate the effect of color, we randomly choose one representative channel for every frame in a tuple, called \textit{channel splitting (Split)}. 
We also explore the other two strategies:
\textit{Swap} randomly swaps two channels, and \textit{Drop} randomly drops one or two channels. 
\figref{dataSampling_colorsplit} shows the gains of using the proposed channel splitting over other alternative strategies.

\vspace{\paramargin}
\paragraph{Pairwise feature extraction.} 
We show the effect of the pairwise feature extraction stage as well as the performance correlation between the sequence sorting task and action recognition. 
We evaluate the order prediction task on a held-out validation set from the automatically sampled data. 
\tabref{dataSampling_suborder} shows the results. 
For both 3-tuple and 4-tuple, models with the pairwise feature extraction perform better than models with simple concatenation on both order prediction and action recognition tasks.
The improvement of the pairwise feature extraction over concatenation is larger on 4-tuple than on 3-tuple due to the increased level of difficulty for the order prediction task.

\begin{figure}[t]
\centering
\subfloat[Action recognition]{%
\includegraphics[width=0.5\columnwidth]{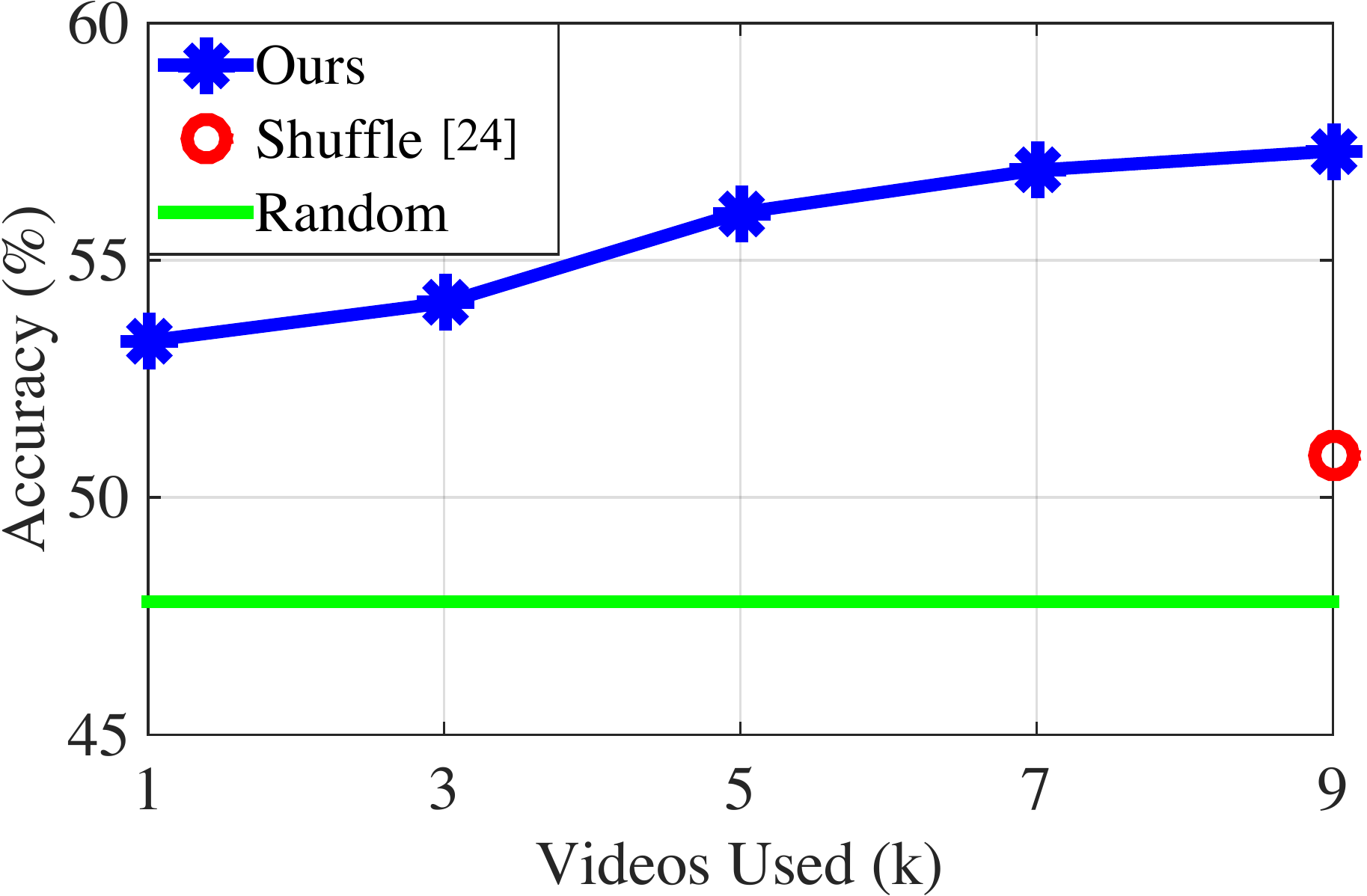}
}
\subfloat[Classification]{%
\includegraphics[width=0.5\columnwidth]{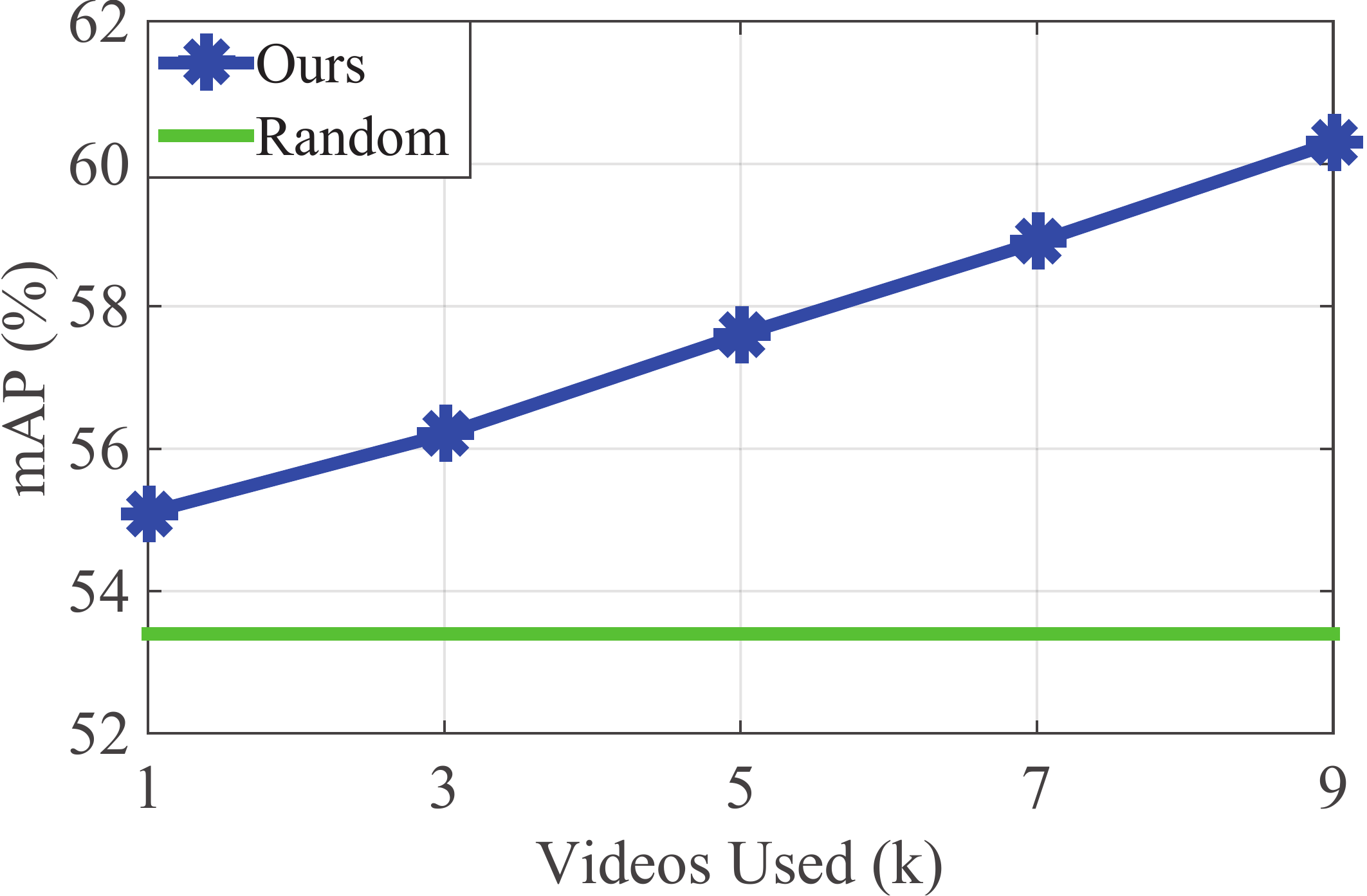}
}
\vspace{\subfigmargin}
\caption{\textbf{Performance comparison using a different amount of videos}. The results show a steady performance improvement when training with more videos. We also show that the unsupervised pre-training offers significant advantages over random initialization.}
\label{figure:ucfvoc}
\vspace{\figmargin}
\end{figure}

\vspace{\paramargin}
\paragraph{Number of training videos.}
We demonstrate the scalability and potential of our method by comparing the performance of using a different amount of videos for unsupervised pre-training. 
\figref{ucfvoc} shows the results on the UCF-101 and the PASCAL VOC 2007 datasets. 
On the UCF-101 dataset, our approach outperforms~\cite{misra2016shuffle} using only 1k videos for pre-training. 
For the classification task on the PASCAL VOC 2007 dataset, the performance consistently improves with the number of training videos.
Training with large-scale and diverse videos~\cite{abu2016youtube} is a promising future direction.
%
%
\input{vis_conv1}
\begin{figure}[t]
\centering
\includegraphics[width=\columnwidth]{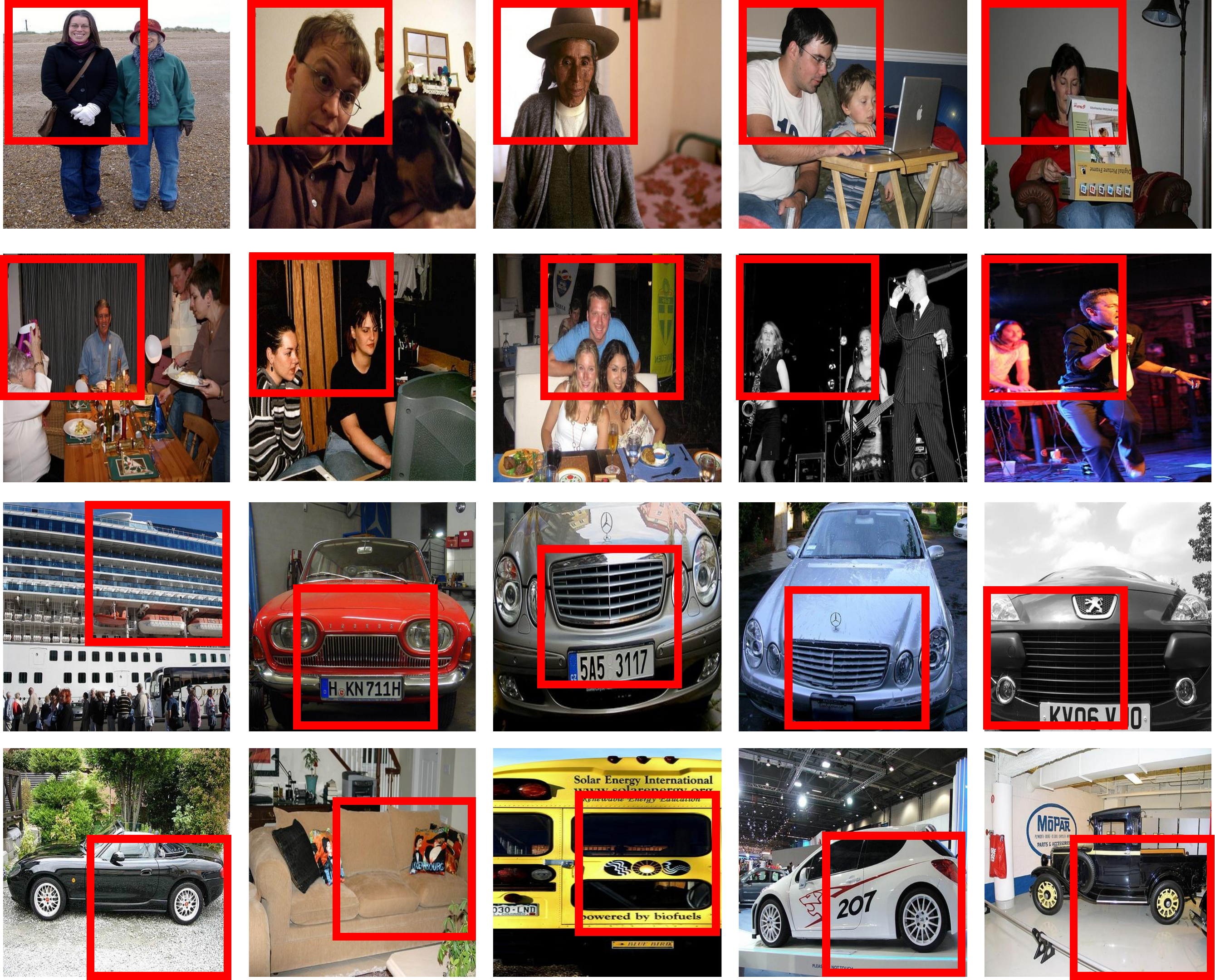}
\vspace{\subfigmargin}
\caption{\textbf{Activation of Pool5 units.} 
Each row lists the top 5 patches that activate a specific unit from the VOC dataset.
While we train the network on the UCF-101 dataset 
without using any manual annotations, 
the pool5 feature activations correspond to human head (\nth{1} and \nth{2} rows) and object parts (\nth{3} and \nth{4} rows).
}
\label{figure:activation}
\vspace{\figmargin}
\end{figure}

\vspace{\secmargin}
\subsection{Visualization}
\label{subsec:vis}
We demonstrate the quality of the learned features by visualizing low-level first layer filter (conv1) as well as high-level activations (pool5). 

\figref{visualization} shows the visualization of the learned filters in conv1. 
\figref{visualization}(a) and (b) show that although using all color channels enable the network to learn some color filters, there are many ``color patch'' filters (see the first two rows in \figref{visualization}(b)). 
These filters lack generalizability and easily make further fine-tuning stuck at a bad initialization.
Comparing \figref{visualization}(c) and (d), 
the filters are sharper and of more varieties when initialized by our method.

\figref{activation} shows the top 5 activations of several pool5 units on the Pascal VOC 2007 dataset.
Although our model is trained on the UCF-101 dataset which focuses on action classes, it captures some meaningful regions without fine-tuning.
For example, the first two rows are human-related and the the third and fourth rows capture the front of cars and wheel-like object, respectively.


\vspace{\secmargin}
\section{Conclusions}

In this paper, we present an unsupervised representation method through solving the sequence sorting problem (sorting a shuffled sequence into a chronological order). 
We propose an Order Prediction Network architecture to facilitate the training. 
Using our approach as pre-training, we demonstrate improved performance over state-of-the-art methods on the UCF-101 and HMDB-51 datasets. 
We also show the competitive generalization ability on classification and detection tasks. 
While promising results have been shown, there is still a performance gap between the unsupervised pre-training and the supervised pre-training methods.
We believe that modeling the long-term evolution in videos (\eg combining with a recurrent neural network) is a promising future direction.

\vspace{\secmargin}
\section*{Acknowledgements}
\vspace{\secmargin}
This work is supported in part by the NSF CAREER Grant \#1149783, gifts from Verisk, Adobe and NVIDIA.

\clearpage

{\small
\bibliographystyle{ieee}
\bibliography{iccv17}
}

\end{document}

%% file: macro.tex
\usepackage{mathptmx}
\usepackage{xspace}
\usepackage{color}
\usepackage{multirow}

\graphicspath{{figures/}}

\newcommand{\tb}[1]{\textbf{#1}}

\long\def\ignorethis#1{}

\definecolor{gray}{rgb}{0.35,0.35,0.35}
\definecolor{MyBlue}{rgb}{0,0.2,0.8}
\definecolor{MyRed}{rgb}{0.8,0.2,0}
\definecolor{MyGreen}{rgb}{0.0,0.5,0.1}
\definecolor{MyGray}{rgb}{0.4,0.4,0.4}

\def\red#1{\textcolor{red}{#1}}
\def\blue#1{\textcolor{blue}{#1}}

\newlength\paramargin
\newlength\figmargin
\newlength\subfigmargin
\newlength\secmargin
\newlength\tabmargin

\usepackage{array}
\newcolumntype{L}[1]{>{\raggedright\let\newline\\\arraybackslash\hspace{0pt}}m{#1}}
\newcolumntype{C}[1]{>{\centering\let\newline\\\arraybackslash\hspace{0pt}}m{#1}}
\newcolumntype{R}[1]{>{\raggedleft\let\newline\\\arraybackslash\hspace{0pt}}m{#1}}

\def\ie{i.e.,~}
\def\eg{e.g.,~}

\def\etal{et~al.\xspace}

\newcommand{\secref}[1]{Section~\ref{sec:#1}}
\newcommand{\subsecref}[1]{Section~\ref{subsec:#1}}
\newcommand{\figref}[1]{Figure~\ref{figure:#1}}
\newcommand{\tabref}[1]{Table~\ref{tab:#1}}

%% file: table_pascal.tex
\begin{table*}[t]
	\caption{\textbf{Results of the Pascal VOC2007 classification and detection datasets.} }
	\label{tab:pascal}
	\vspace{-2mm}
	\centering
	\small
	\begin{tabular}{lccccc} \toprule
		Method & Pretraining time & Source & Supervision & Classification & Detection \\
		\midrule
		Krizhevsky \etal \cite{krizhevsky2012imagenet} & 3 days & ImageNet & labeled classes & 78.2 & 56.8 \\
		\midrule
		
		Doerch \etal \cite{doersch2015context} & 4 weeks & ImageNet & context & 55.3 & 46.6 \\
		Pathak \etal \cite{pathak2016inpainting} & 14 hours & ImagetNet+StreetView & context & 56.5 & 44.5  \\
		Norrozi \etal \cite{noroozi2016puzzle} & 2.5 days & ImageNet & context & \textbf{\red{68.6}} & \textbf{\red{51.8}} \\ 
		Zhang \etal \cite{zhang2016splitbrain} & - & ImageNet & reconstruction & \blue{\underline{67.1}} & \blue{\underline{46.7}} \\
		\midrule
		Wang and Gupta (color) \cite{wang2015unsupervised} & 1 weeks & 100k videos, VOC2012 & motion & 58.4 & 44.0 \\
		Wang and Gupta (grayscale) \cite{wang2015unsupervised} & 1 weeks & 100k videos, VOC2012 & motion & \blue{\underline{62.8}} & \textbf{\red{47.4}} \\
		Agrawal \etal \cite {agrawal2015ego} & - &　KITTI, SF & motion & 52.9 & 41.8 \\
		Misra \etal \cite{misra2016shuffle} & - & $<$ 10k videos  & motion & 54.3 & 39.9 \\
		Ours (OPN) & $<$ 3 days & $<$ 30k videos  & motion & \textbf{\red{63.8}} & \blue{\underline{46.9}} \\
		\bottomrule
	\end{tabular}
	\vspace{-2mm}
\end{table*}

%% file: vis_conv1.tex
\begin{figure}[t]
	\centering
	\captionsetup[subfloat]{farskip=2pt,captionskip=1pt}
	\subfloat[With channel splitting]{%
		\includegraphics[width=0.49\linewidth]{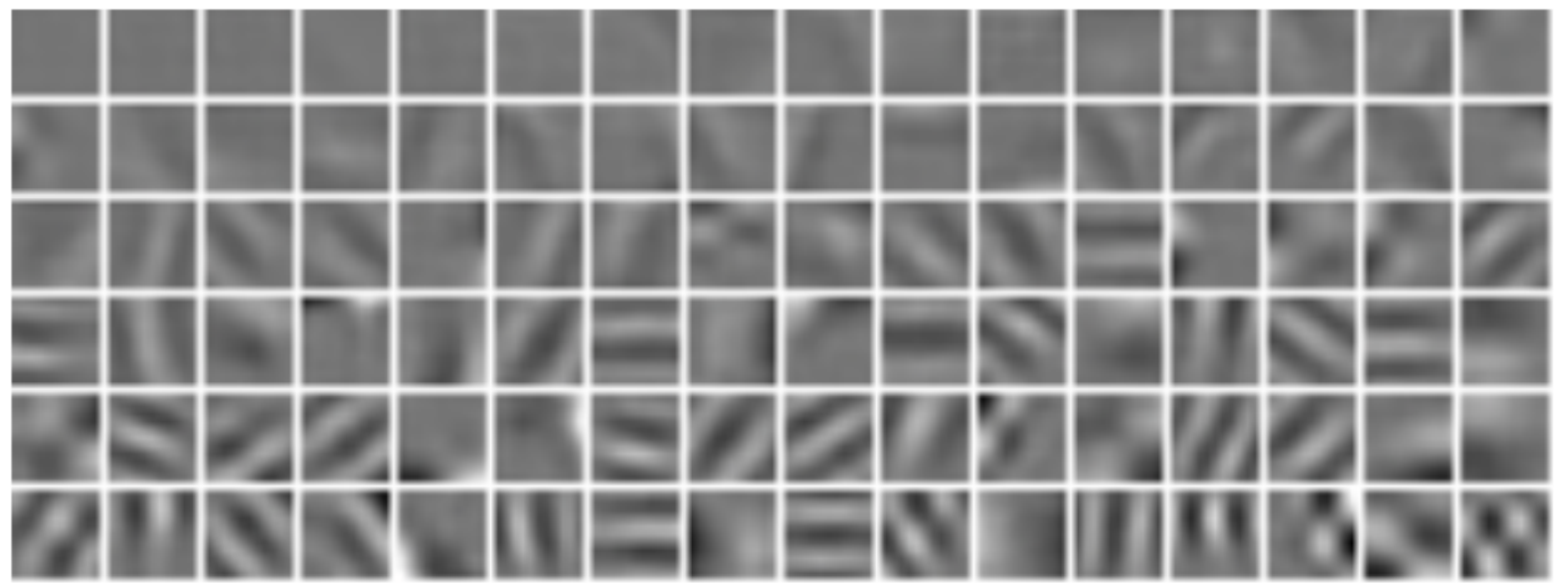}%
	}\hfill
	\subfloat[With RGB frames]{%
		\includegraphics[width=0.49\linewidth]{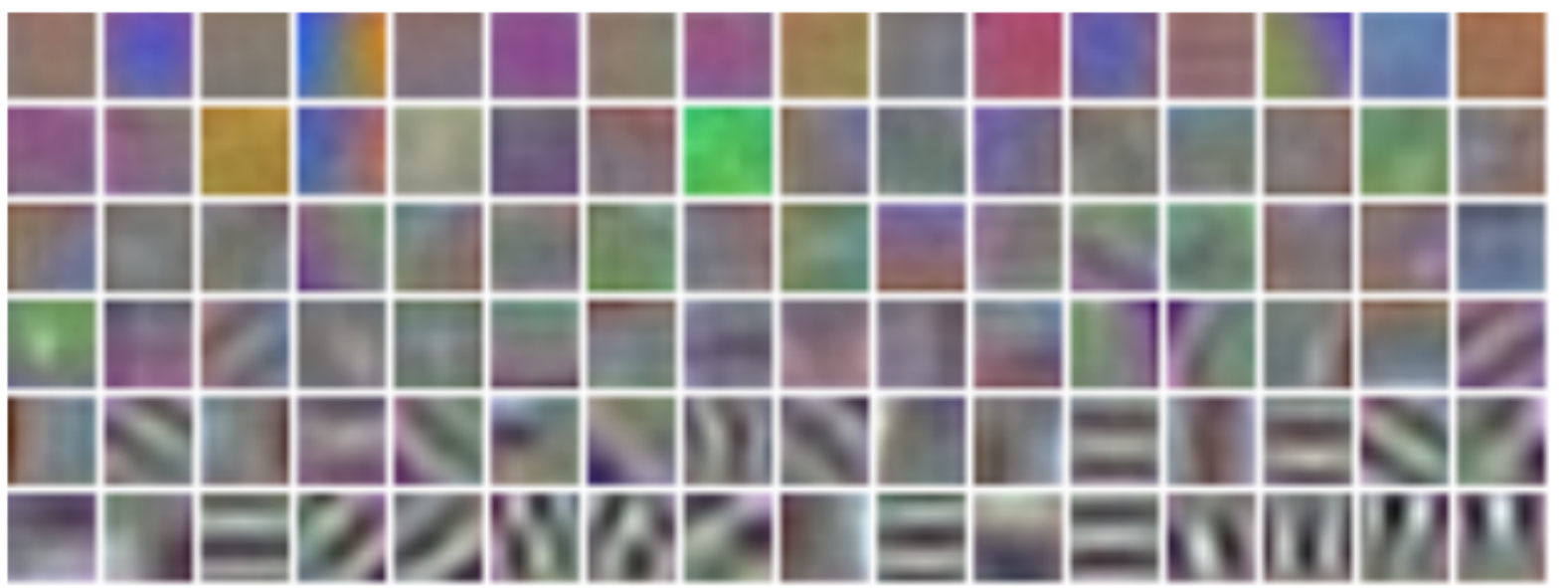}%
	}
	\vspace{1mm}
	\subfloat[With channel splitting, fine-tuned on UCF-101]{%
		\includegraphics[width=0.49\linewidth]{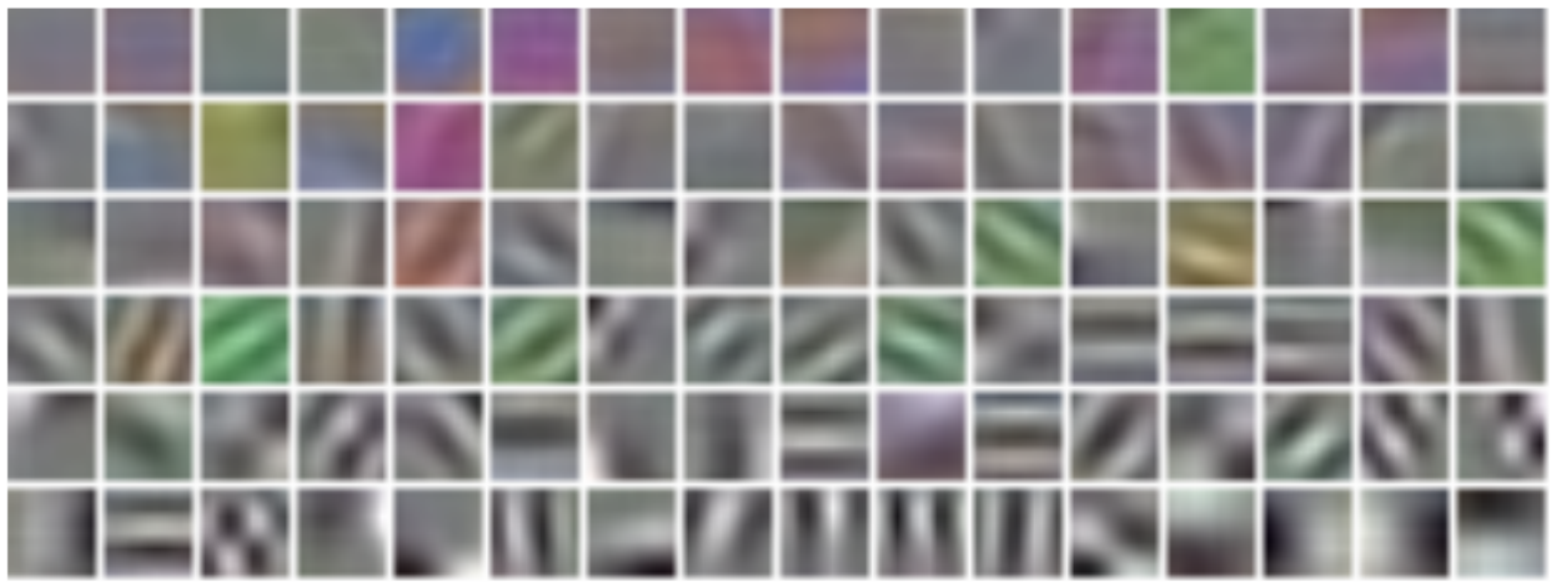}%
	} \hfill
	\subfloat[Random initialization, trained on UCF-101]{%
		\includegraphics[width=0.49\linewidth]{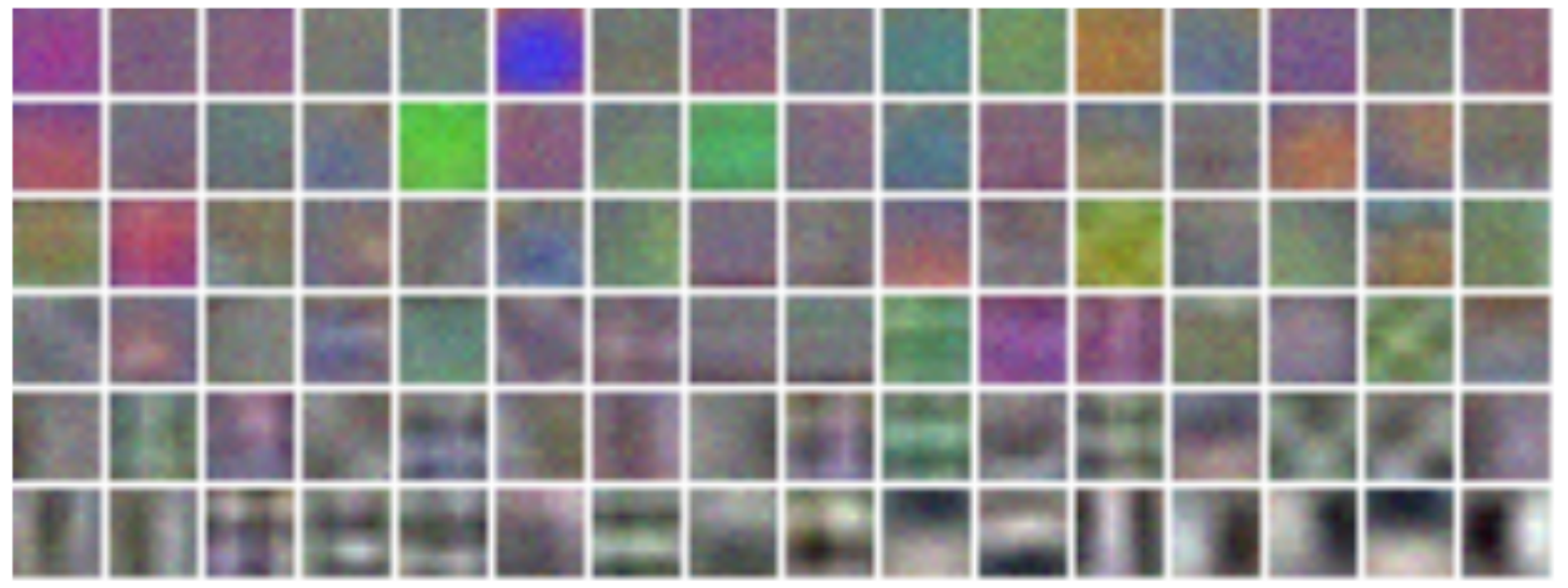}%
	}
	\vspace{-2mm}
	\caption{\textbf{Visualization of Conv1 filters.} Filters in (a)(b) are trained on the UCF-101 dataset in an unsupervised manner.
		Filters in (c) are fine-tuned from filters in (a) on the UCF-101 dataset with supervision, while filters in (d) are trained from scratch. 
		Note that those ``color patch'' filters are usually not desirable because they tend to make the further fine-tuning stuck at a bad initialization.
	}
	\label{figure:visualization}
	\vspace{-0.3cm}
\end{figure}

%% file: iccv17_OPN.bbl
\begin{thebibliography}{10}\itemsep=-1pt

\bibitem{abu2016youtube}
S.~Abu-El-Haija, N.~Kothari, J.~Lee, P.~Natsev, G.~Toderici, B.~Varadarajan,
  and S.~Vijayanarasimhan.
\newblock Youtube-8m: A large-scale video classification benchmark.
\newblock {\em arXiv preprint arXiv:1609.08675}, 2016.

\bibitem{agrawal2015ego}
P.~Agrawal, J.~Carreira, and J.~Malik.
\newblock Learning to see by moving.
\newblock In {\em ICCV}, 2015.

\bibitem{bengio2007stackingRBMandAutoencoder}
Y.~Bengio, P.~Lamblin, D.~Popovici, H.~Larochelle, et~al.
\newblock Greedy layer-wise training of deep networks.
\newblock In {\em NIPS}, 2007.

\bibitem{deng2009imagenet}
J.~Deng, W.~Dong, R.~Socher, L.-J. Li, K.~Li, and L.~Fei-Fei.
\newblock Imagenet: A large-scale hierarchical image database.
\newblock In {\em CVPR}, 2009.

\bibitem{doersch2013midlevel2}
C.~Doersch, A.~Gupta, and A.~A. Efros.
\newblock Mid-level visual element discovery as discriminative mode seeking.
\newblock In {\em NIPS}, 2013.

\bibitem{doersch2015context}
C.~Doersch, A.~Gupta, and A.~A. Efros.
\newblock Unsupervised visual representation learning by context prediction.
\newblock In {\em ICCV}, 2015.

\bibitem{everingham2010pascal}
M.~Everingham, L.~Van~Gool, C.~K. Williams, J.~Winn, and A.~Zisserman.
\newblock The pascal visual object classes (voc) challenge.
\newblock {\em IJCV}, 88(2):303--338, 2010.

\bibitem{fernando2016O3N}
B.~Fernando, H.~Bilen, E.~Gavves, and S.~Gould.
\newblock Self-supervised video representation learning with odd-one-out
  networks.
\newblock In {\em CVPR}, 2017.

\bibitem{girshick2015fastrcnn}
R.~Girshick.
\newblock Fast r-cnn.
\newblock In {\em ICCV}, 2015.

\bibitem{hinton2006stackingRBM}
G.~E. Hinton and R.~R. Salakhutdinov.
\newblock Reducing the dimensionality of data with neural networks.
\newblock {\em Science}, 313(5786):504--507, 2006.

\bibitem{ioffe2015batchnorm}
S.~Ioffe and C.~Szegedy.
\newblock Batch normalization: Accelerating deep network training by reducing
  internal covariate shift.
\newblock In {\em ICML}, 2015.

\bibitem{isola2015cooccur}
P.~Isola, D.~Zoran, D.~Krishnan, and E.~H. Adelson.
\newblock Learning visual groups from co-occurrences in space and time.
\newblock In {\em ICLR, Workshop}, 2016.

\bibitem{jayaraman2015ego}
D.~Jayaraman and K.~Grauman.
\newblock Learning image representations tied to ego-motion.
\newblock In {\em ICCV}, 2015.

\bibitem{jayaraman2015slowandsteady}
D.~Jayaraman and K.~Grauman.
\newblock Slow and steady feature analysis: Higher order temporal coherence in
  video.
\newblock In {\em CVPR}, 2016.

\bibitem{jia2014caffe}
Y.~Jia, E.~Shelhamer, J.~Donahue, S.~Karayev, J.~Long, R.~Girshick,
  S.~Guadarrama, and T.~Darrell.
\newblock Caffe: Convolutional architecture for fast feature embedding.
\newblock In {\em ACM MM}, 2014.

\bibitem{krahenbuhl2015datadependent}
P.~Kr{\"a}henb{\"u}hl, C.~Doersch, J.~Donahue, and T.~Darrell.
\newblock Data-dependent initializations of convolutional neural networks.
\newblock In {\em ICLR}, 2015.

\bibitem{krizhevsky2012imagenet}
A.~Krizhevsky, I.~Sutskever, and G.~E. Hinton.
\newblock Imagenet classification with deep convolutional neural networks.
\newblock In {\em NIPS}, 2012.

\bibitem{kuehne2011hmdb}
H.~Kuehne, H.~Jhuang, E.~Garrote, T.~Poggio, and T.~Serre.
\newblock Hmdb: a large video database for human motion recognition.
\newblock In {\em ICCV}, 2011.

\bibitem{larsson2016colorization}
G.~Larsson, M.~Maire, and G.~Shakhnarovich.
\newblock Learning representations for automatic colorization.
\newblock In {\em ECCV}, 2016.

\bibitem{le2013autoendocerLargescale}
Q.~V. Le.
\newblock Building high-level features using large scale unsupervised learning.
\newblock In {\em ICML}, 2012.

\bibitem{li2016unsupervised}
D.~Li, W.-C. Hung, J.-B. Huang, S.~Wang, N.~Ahuja, and M.-H. Yang.
\newblock Unsupervised visual representation learning by graph-based consistent
  constraints.
\newblock In {\em ECCV}, 2016.

\bibitem{long2016interpolation}
G.~Long, L.~Kneip, J.~M. Alvarez, H.~Li, X.~Zhang, and Q.~Yu.
\newblock Learning image matching by simply watching video.
\newblock In {\em ECCV}, 2016.

\bibitem{lotter2017Prenet}
W.~Lotter, G.~Kreiman, and D.~Cox.
\newblock Deep predictive coding networks for video prediction and unsupervised
  learning.
\newblock In {\em ICLR}, 2017.

\bibitem{misra2016shuffle}
I.~Misra, C.~L. Zitnick, and M.~Hebert.
\newblock Shuffle and learn: Unsupervised learning using temporal order
  verification.
\newblock In {\em ECCV}, 2016.

\bibitem{mobahi2009temporalcoherence}
H.~Mobahi, R.~Collobert, and J.~Weston.
\newblock Deep learning from temporal coherence in video.
\newblock In {\em ICML}, 2009.

\bibitem{noroozi2016puzzle}
M.~Noroozi and P.~Favaro.
\newblock Unsupervised learning of visual representations by solving jigsaw
  puzzles.
\newblock In {\em ECCV}, 2016.

\bibitem{olshausen1997oldautoencoder}
B.~A. Olshausen and D.~J. Field.
\newblock Sparse coding with an overcomplete basis set: A strategy employed by
  v1?
\newblock {\em Vision research}, 37(23):3311--3325, 1997.

\bibitem{owens2016ambient}
A.~Owens, J.~Wu, J.~H. McDermott, W.~T. Freeman, and A.~Torralba.
\newblock Ambient sound provides supervision for visual learning.
\newblock In {\em ECCV}, 2016.

\bibitem{pathak2016inpainting}
D.~Pathak, P.~Kr{\"a}henb{\"u}hl, J.~Donahue, T.~Darrell, and A.~A. Efros.
\newblock Context encoders: Feature learning by inpainting.
\newblock In {\em CVPR}, 2016.

\bibitem{purushwalkam2016transformation}
S.~Purushwalkam and A.~Gupta.
\newblock Pose from action: Unsupervised learning of pose features based on
  motion.
\newblock In {\em ECCV, Workshop}, 2016.

\bibitem{raina2007self}
R.~Raina, A.~Battle, H.~Lee, B.~Packer, and A.~Y. Ng.
\newblock Self-taught learning: Transfer learning from unlabeled data.
\newblock In {\em ICML}, 2007.

\bibitem{russell2006handcraft1}
B.~C. Russell, W.~T. Freeman, A.~A. Efros, J.~Sivic, and A.~Zisserman.
\newblock Using multiple segmentations to discover objects and their extent in
  image collections.
\newblock In {\em CVPR}, 2006.

\bibitem{simonyan2014VGG}
K.~Simonyan and A.~Zisserman.
\newblock Very deep convolutional networks for large-scale image recognition.
\newblock {\em CoRR}, abs/1409.1556, 2014.

\bibitem{singh2012midlevel1}
S.~Singh, A.~Gupta, and A.~A. Efros.
\newblock Unsupervised discovery of mid-level discriminative patches.
\newblock In {\em ECCV}, 2012.

\bibitem{sivic2005handcraft2}
J.~Sivic, B.~C. Russell, A.~A. Efros, A.~Zisserman, and W.~T. Freeman.
\newblock Discovering objects and their location in images.
\newblock In {\em ICCV}, 2005.

\bibitem{soomro2012ucf101}
K.~Soomro, A.~R. Zamir, and M.~Shah.
\newblock U{C}{F}101: A dataset of 101 human actions classes from videos in the
  wild.
\newblock {\em arXiv preprint arXiv:1212.0402}, 2012.

\bibitem{srivastava2015LSTM}
N.~Srivastava, E.~Mansimov, and R.~Salakhutdinov.
\newblock Unsupervised learning of video representations using lstms.
\newblock In {\em ICML}, 2015.

\bibitem{sun2013midlevel3}
J.~Sun and J.~Ponce.
\newblock Learning discriminative part detectors for image classification and
  cosegmentation.
\newblock In {\em ICCV}, 2013.

\bibitem{vondrick2016generating}
C.~Vondrick, H.~Pirsiavash, and A.~Torralba.
\newblock Generating videos with scene dynamics.
\newblock In {\em NIPS}, 2016.

\bibitem{wang2016act}
X.~Wang, A.~Farhadi, and A.~Gupta.
\newblock Actions {\textasciitilde} transformations.
\newblock In {\em CVPR}, 2016.

\bibitem{wang2015unsupervised}
X.~Wang and A.~Gupta.
\newblock Unsupervised learning of visual representations using videos.
\newblock In {\em CVPR}, 2015.

\bibitem{zhang2016colorization2}
R.~Zhang, P.~Isola, and A.~A. Efros.
\newblock Colorful image colorization.
\newblock In {\em ECCV}, 2016.

\bibitem{zhang2016splitbrain}
R.~Zhang, P.~Isola, and A.~A. Efros.
\newblock Split-brain autoencoders: Unsupervised learning by cross-channel
  prediction.
\newblock In {\em CVPR}, 2017.

\end{thebibliography}
